\documentclass[a4paper, 10pt, conference]{ieeeconf}      
\usepackage{FG2025}

\FGfinalcopy 

\IEEEoverridecommandlockouts                              
\usepackage{multirow} 
\usepackage{epsfig} 
\usepackage{mathptmx} 
\usepackage{times} 
\usepackage{amsmath} 
\usepackage{amssymb}  
\usepackage{placeins}
\def\FGPaperID{7} 

\title{\LARGE \bf
FreqMixFormerV2: Lightweight Frequency-aware Mixed Transformer for Human Skeleton Action Recognition
}

\author{\parbox{16cm}{\centering
    {\large Wenhan Wu$^1$, Pengfei Wang$^2$, Chen Chen$^3$, Aidong Lu$^1$}\\
    {\normalsize
    $^1$ Department of Computer Science, University of North Carolina at Charlotte, Charlotte, USA\\
    $^2$ Independent Researcher\\
    $^3$ Center for Research in Computer Vision, University of Central Florida, Orlando, USA
    }}
}

\usepackage{wrapfig}
\usepackage{graphicx}
\usepackage{pifont}
\usepackage{amssymb}
\usepackage{amsmath}
\newcommand{\Checkmark}{\ding{51}} 
\newcommand{\XSolidBrush}{\ding{55}}
\usepackage{float}
\usepackage{url}
\usepackage{multicol}
\usepackage{hyperref}
\usepackage{colortbl} 
\definecolor{lightgray}{gray}{0.9} 
\begin{document}

\ifFGfinal
\thispagestyle{empty}
\pagestyle{empty}
\else
\author{Anonymous FG2025 submission\\ Paper ID \FGPaperID \\}
\pagestyle{plain}
\fi
\maketitle

\begin{abstract}



Transformer-based human skeleton action recognition has been developed for years. However, the complexity and high parameter count demands of these models hinder their practical applications, especially in resource-constrained environments. 
In this work, we propose FreqMixForemrV2, which was built upon the Frequency-aware Mixed Transformer (FreqMixFormer) \cite{wu2024frequency} for identifying subtle and discriminative actions with pioneered frequency-domain analysis.
We design a lightweight architecture that maintains robust performance while significantly reducing the model complexity. This is achieved through a redesigned frequency operator that optimizes high-frequency and low-frequency parameter adjustments,  and a simplified frequency-aware attention module. 
These improvements result in a substantial reduction in model parameters, enabling efficient deployment with only a minimal sacrifice in accuracy.
Comprehensive evaluations of standard datasets (NTU RGB+D, NTU RGB+D 120, and NW-UCLA datasets) demonstrate that the proposed model achieves a superior balance between efficiency and accuracy, outperforming state-of-the-art methods with only 60\% of the parameters. Our project is publicly available at: \href{https://github.com/wenhanwu95/FreqMixFormer}{\textcolor{magenta}{https://github.com/wenhanwu95/FreqMixFormer}}.

\end{abstract}

\section{INTRODUCTION}

The continuous development of transformer \cite{vaswani2017attention} has achieved significant advancements in skeletal action recognition \cite{shi2020decoupled, plizzari2021skeleton, zhang2021stst, bai2022hierarchical, gao2022focal, zhou2022hypergraph, liu2023transkeleton, xin2023skeleton, wu2023skeletonmae, wu2024frequency}. Transformer-based models have emerged as powerful tools to capture complex spatial and temporal relationships within skeletal data. These models excel at recognizing actions by capturing the sequential dependencies of skeletal joints, enabling them to distinguish intricate human motions. 

As the pioneering work among transformer-based methods, FreqMixFormer \cite{wu2024frequency} leverages frequency-domain analysis to enhance action recognition. By focusing on frequency components, FreqMixFormer captures the discriminative details of skeletal movements that traditional spatial and temporal models often overlook. This allows the model to distinguish the confusing actions, improving recognition accuracy. However, despite its effectiveness in recognition, the architecture of FreqMixFormer is inherently complex and computationally intensive. The high parameter count and significant computational overhead are challenging for real-world deployment, particularly when resources are limited or real-time processing is required.

\begin{figure}[htp]
\vspace{-5pt}
  \centering
  \includegraphics[width=1.0\linewidth]{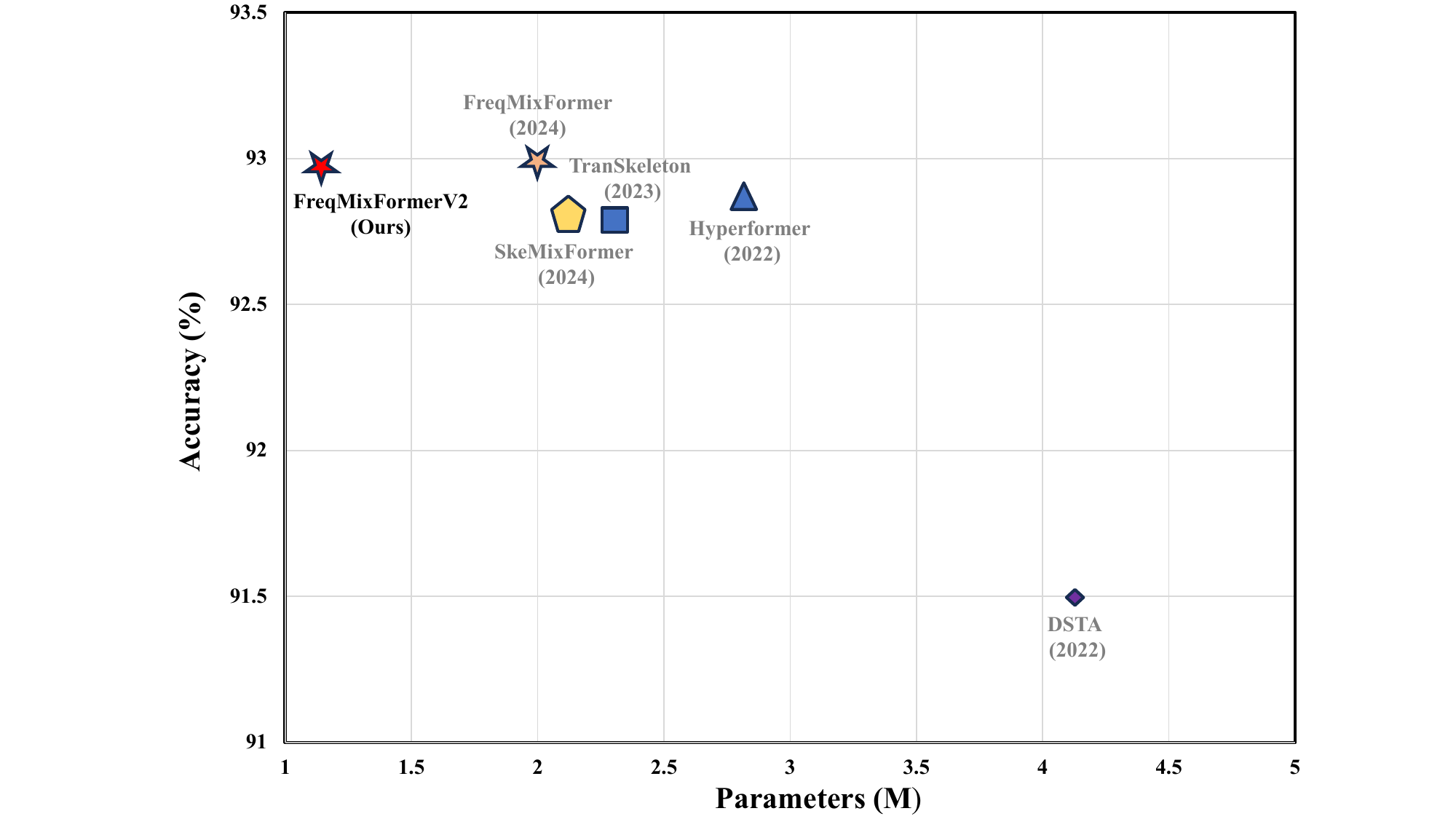}
  \vspace{-15pt}
  \caption{Performance vs. model size on NTU-60 \cite{shahroudy2016ntu} X-Sub setting. Our FreqMixFormerV2 demonstrates superior performance and efficiency compared to previous transformer-based methods. Notably, our method reduces the number of parameters by nearly half compared to FreqMixFormer while still achieving state-of-the-art accuracy.
}
  \label{fig:fig1}
  \vspace{-15pt}
\end{figure}

Two factors limit the efficiency of the FreqMixFormer. (a) The design of mixed attention blocks for both spatial and frequency attention modules, which captures the contextual information in a mixed way by calculating the self-attention map (\(Q_i \) and \(K_i \)) and the mix-attention map (\(Q_{i+1} \) and \(K_i \), \(Q_i \) and \(K_{i-1} \)). However, densely applying (e.g., the number of mixed attention blocks, $n$ = 4 in the paper) these modules is computationally expensive with a large parameter count (e.g., 2.0M). (b) Design of the frequency operator. This module is designed for frequency coefficient fine-tuning with a frequency operator $\varphi$. However, the FreqMixFormer ignores the analysis of how high and low frequencies are delineated. 

In this work, we address these challenges by proposing a refined version of the frequency-aware mixed transformer that significantly reduces the computational burden while maintaining strong performance, named \textbf{FreqMixFormerV2}. First, we simplify the mixed Frequency-aware Mixed Transformer by combining the high-frequency attention block (HFAB), low-frequency attention block (LFAB), and Spatial attention block (SAB). This allows the model to retain its ability to capture fine-grained details in skeletal movements while reducing the computational cost. Second, a new high-low-frequency operator is proposed to replace the original one that ignores the high-low frequency difference, enabling it to modulate high-frequency and low-frequency coefficients more efficiently. Extensive experiments on three widely used benchmarks show that our approach reduces the parameter count by half compared to the original method while maintaining comparable performance. The main contributions are:

\begin{itemize} 
\item We propose a simplified and optimized architecture by introducing distinct high-frequency (HFAB) and low-frequency attention blocks (LFAB) along with a spatial attention block, which reduces the computational complexity significantly the computational complexity while maintaining the model’s capability.

\item We introduce a new high-low-frequency operator, which efficiently modulates frequency coefficients. This enhancement optimizes the frequency domain analysis, improving the model's ability to adjust high-frequency and low-frequency components and increasing the ability of the frequency operator.

\item Extensive experiments on three datasets (NTU RGB+D \cite{shahroudy2016ntu}, NTU RGB+D 120 \cite{liu2019ntu}, and Northwestern-UCLA \cite{wang2014cross}) show that FreqMixFormerV2 achieves comparable performance with only a 0.8\% reduction in accuracy, while using just 60\% of the parameter size of the original model.
\end{itemize}

\section{METHOD}
\subsection{FreqMixFormerV2 vs. FreqMixFormer}
\label{sec:vs}
\begin{figure*}[htp]
\vspace{-15pt}
  \centering
  \includegraphics[width=1\linewidth]{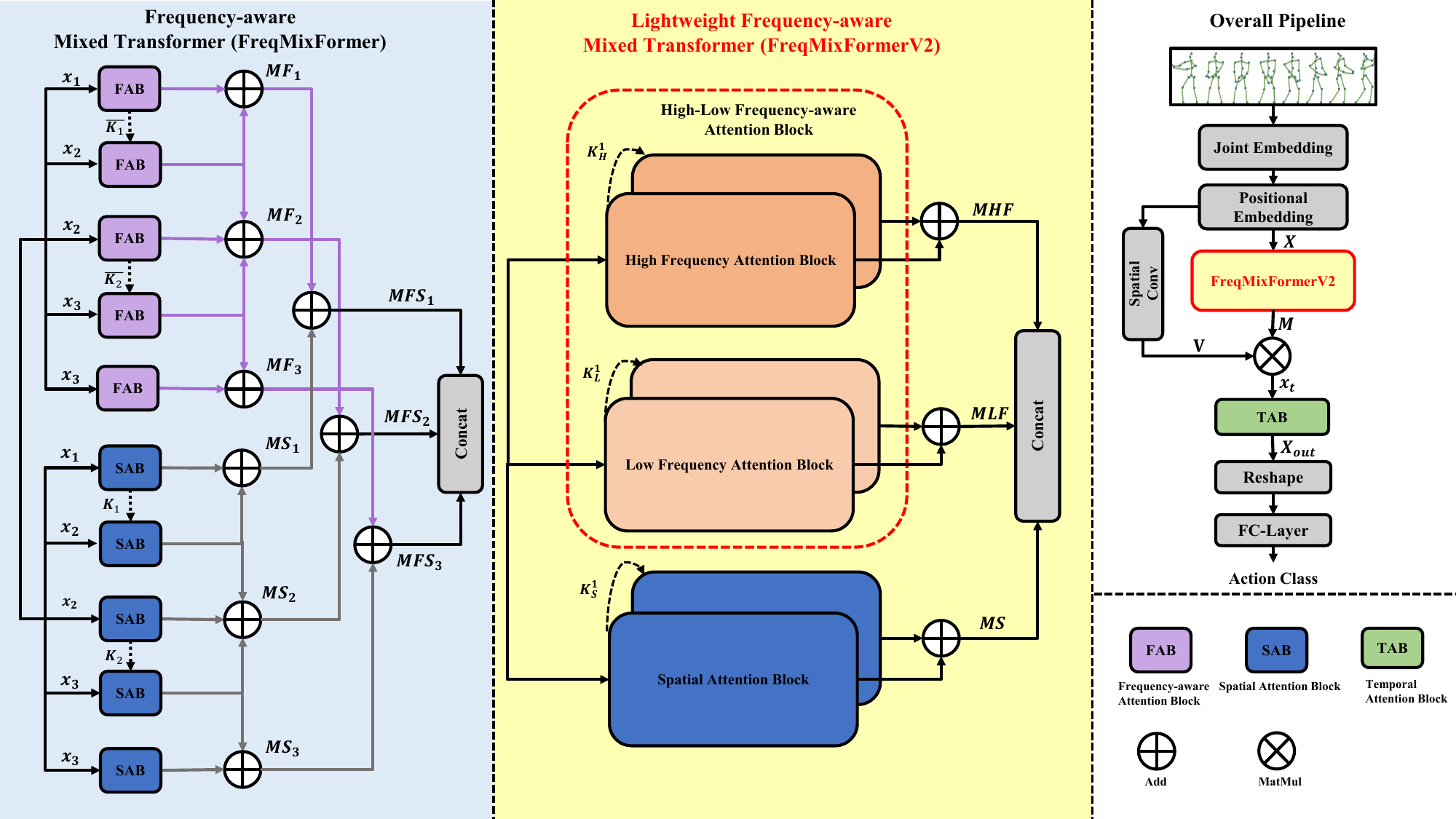}
  \vspace{-15pt}
  \caption{\small{FreqMixFormerV2 vs. FreqMixFormer. 
Given the skeleton sequence, we first apply joint and positional embeddings to obtain the embedded representation, denoted as $X$. Then $X$ is processed by a high-low frequency-aware attention block (combined with high-frequency and low-frequency attention blocks, as shown in Fig. \ref{fig:fig3}) to extract a mixed high-low frequency attention map. A spatial attention block is also employed for mixed spatial attention maps. These maps are then concatenated into a feature $M$, along with the \textit{Value} $V$, to serve as the input for the temporal attention block. This process facilitates inter-frame joint correlation learning and the resulting output, $X_{out}$ for action classification. The difference analysis can be found in Section \ref{sec:vs}}.}
  \label{fig:fig2}
  \vspace{-5pt}
\end{figure*}

As shown in Fig.\ref{fig:fig2}, FreqMixFormerV2 follows the overall structure of FreqMixFormer, but \textbf{contains three key differences which lead to its lightweight and efficient performance:}
1) \textbf{Simplify the structure:}  
Unlike the 4-unit inputs of FreqMixFormer for multi-channel inputs \(x_i\), FreqMixFormerV2 simplifies the input to only 2 channels in the data processing. This reduces the computational load while retaining essential skeletal information.
2) \textbf{Lightweight Frequency-aware Mixed Transformer Design:}  
The Frequency-aware Mixed Transformer in FreqMixFormerV2 is streamlined with fewer modules (2 blocks for each module vs. 4 blocks for each module, e.g., the number of SAB blocks) and parameters (1.2M vs. 2.0M, illustrated in Fig.\ref{fig:fig1}). This lightweight design preserves spatial and frequency interactions but with reduced parameter usage, making it more efficient.
3) \textbf{Improved High-Low Frequency Operator:}  
FreqMixFormerV2 enhances frequency feature learning by dynamically amplifying high-frequency and down-weighting low-frequency components. This improves recognition performance without adding complexity. A comparison table is listed in the Table. \ref{table vs} for motivation clarification.

\vspace{-5pt}
\begin{table}[H]
\renewcommand\arraystretch{1.5}
\centering
\caption{Comparison between FreqMixFormer and FreqMixFormerV2}
\resizebox{\linewidth}{!}{
\begin{tabular}{|l|c|c|}
\hline
\textbf{Feature} & \textbf{FreqMixFormer} & \textbf{FreqMixFormerV2 (ours)} \\ \hline
\textbf{Input Channels} & Multi-channel, the number of input channel, \( n = 4 \)  & 2-channel, \( n = 2 \) \\ \hline
\textbf{Transformer Modules} & 7 FAB, 7 SAB, and 1 TAB (15 in total) & 2 HFAB, 2LFAB, 1 SAB and 1 TAB (6 in total)\\ \hline
\textbf{Frequency Operator} & Uniform Frequency Operator (only 1 operator)& High-Low Frequency Operator (2 operators)\\ \hline
\textbf{Parameter Count} & High (2M) & Significantly reduced (1.2M) \\ \hline
\textbf{Performance} & State-of-the-art (91.5 on NTU-60 X-Sub) & Comparable (90.7 on NTU-60 X-Sub)\\ \hline
\end{tabular}
}
\label{table vs}
\end{table}

\vspace{-12pt}
\subsection{Data Processing}
\label{data processing}
Given the input \( X \in \mathbb{R}^{J \times C \times F} \), representing a skeleton sequence of \(F\) frames, where \(C\) is the joint dimensionality and \(J\) is the number of joints per frame, the sequence is embedded using joint and positional embedding layers. Then we divide \(X\) into 2 unit groups through channel splitting based on the data processing from FreqMixformer\cite{wu2024frequency}. The split units are represented as \( x_i \in \mathbb{R}^{J \times (C/2) \times F} \), and the original \(X\) is reconstructed as \( X \leftarrow \text{Concat}[x_1, x_2] \). Each unit \(x_i\) is then processed through the Lightweight Frequency-aware Mixed Transformer, leveraging self-attention and mix-attention mechanisms in both frequency and spatial domains.

\subsection{Lightweight Frequency-aware Mixed Transformer}
\label{sec:FreqMixFormerV2}
\subsubsection{Mixed Spatial Attention}

Given a split input \( x_i \in \mathbb{R}^{J \times (C/2) \times F } \) where \( i = 1, 2 \), the basic Query matrix and Key matrix for each sequence are extracted for spatial feature extraction:
\begin{align}
\label{eq:QK}
\small
  Q_i, K_i = \text{ReLU} (\text{linear}(\text{AvgPool}(x_i))),
\end{align}
where the \(AvgPool\) denotes adaptive average pooling, which smooths the joint weights and minimizes the influence of noisy or less relevant variations in the skeletal data \cite{xin2023skeleton}. An FC-layer with a ReLU activation is applied to ensure that \(Q_i\) and \(K_i\) are globally integrated to get the self-attention map:
\vspace{-3mm}
\begin{align}
\small
  Atten_{self} = Softmax\left(\frac{Q_1 K_1^\top}{\sqrt{d}}\right)
\end{align}

To enable richer contextual integration across the two unit groups, a cross-attention strategy is proposed where \(K_i\) is shared between the attention blocks. The cross-attention is expressed as:

\vspace{-5mm}
\begin{align}
\small 
  Atten_{mix} = Softmax\left(\frac{Q_2 K_1^\top}{\sqrt{d}}\right)
\end{align}

Each mixed spatial attention map is then formulated as:
\begin{align}
\small 
  MS = Atten_{self} + Atten_{mix}
\end{align}

\subsubsection{High-Low Frequency-aware Attention}
\label{sec:high-low}
We apply DCT (the detailed information can be found in the \textcolor{magenta}{Appendix}) to obtain the corresponding frequency coefficients from the split joint sequence \(x_i\) (where \(i = 1, 2\)), and the inputs can be denoted as \(DCT_H(x_i)\) and \(DCT_L(x_i)\), where \(DCT_H(\cdot)\) indicates the input is processed by DCT transform with the high-frequency operator, and \(DCT_L(\cdot)\) indicates the input processed by DCT transform with the low-frequency operator. Similar to the mixed spatial attention, we obtain the Query and Key values along the frequency domain: 
\begin{align}
\small 
  Q_H^i, K_H^i = ReLU (linear(AvgPool(DCT(x_H^i))))
\end{align} 
\vspace{-5mm}
\begin{align}
\small 
  Q_L^i, K_L^i = ReLU (linear(AvgPool(DCT(x_L^i))))
\end{align}

According to the inputs, the high and low frequency-based self-attention and mixed-attention maps can be expressed as:  
\begin{scriptsize}
\begin{align}
    &\begin{aligned}
        Atten_H^{self} &= \text{Softmax}\left(\frac{Q_H^1 K_H^1{}^\top}{\sqrt{d}}\right), &
        Atten_H^{mix} &= \text{Softmax}\left(\frac{Q_H^2 K_H^1{}^\top}{\sqrt{d}}\right), \\
        Atten_L^{self} &= \text{Softmax}\left(\frac{Q_L^1 K_L^1{}^\top}{\sqrt{d}}\right), &
        Atten_L^{mix} &= \text{Softmax}\left(\frac{Q_L^2 K_L^1{}^\top}{\sqrt{d}}\right)
    \end{aligned}
\end{align}
\end{scriptsize}

Thus, the final frequency attention maps are expressed as: 
\begin{align}
\small 
    MHF = Atten_H^{self} + Atten_H^{mix} 
\end{align}
\begin{align}
\small 
    MLF = Atten_L^{self} + Atten_L^{mix} 
\end{align}

Notably, it is different from FreqMixFormer, which simply adopts a Frequency Operator (FO) to mixed frequency attention maps. We introduce a frequency division coefficient \(N\), where \(N \in (1, J)\) and \(J\) are the number of joints, to partition high-frequency and low-frequency parameters: the first \(N\) coefficients are considered low-frequency, while the remaining \(J-N\) coefficients are regarded as high-frequency parameters based on the fundamental definition of DCT, where the initial coefficients correspond to the low-frequency components, and the later coefficients represent the higher-frequency components. Subsequently, a high-frequency operator coefficient \(h\) and a low-frequency operator coefficient \(\ell\) are proposed to fine-tune the DCT coefficient based on their frequency level, where \(\ell \in (0, 1)\) and \(h \in (1, 1+\ell)\). \(h\) is applied to amplify the high-frequency features, emphasizing subtle actions. Conversely, the low-frequency coefficients are attenuated by \(\ell\), reducing the emphasis on salient actions while maintaining the overall integrity of the action representations. The final output of the mixed frequency-spatial attention map can be expressed as:
\begin{align}
\label{eq:FSM}
\small 
      M = \text{Concat}[MS, MHF, MLF]
\end{align} 
The Value \(V\) is derived from the initial input \(X\) by applying a unified computation through a \(1 \times 1\) convolutional layer along the spatial dimension. As a result, the input to the Temporal Attention Block is given by:
\begin{align}
\small 
      x_t = MV
\end{align} 

\begin{table*}[]
\vspace{-15pt}
\scriptsize
\renewcommand\arraystretch{1.2}
\centering
  \caption{ Comparison with the SOTA. J represents joint only, 4S indicates the results are produced with 4 modalities \cite{wu2024frequency}.}
   \setlength\tabcolsep{6.0pt} 
{
\begin{tabular}{cccccccccc}
\hline
\multirow{2}{*}{Methods} &
\multirow{2}{*}{Venue} &
\multirow{2}{*}{Category} &
\multirow{2}{*}{Modalities} &
\multirow{2}{*}{Parameters(M)} &
\multicolumn{2}{c}{NTU RGB+D 60} &
\multicolumn{2}{c}{NTU RGB+D 120} &
\multirow{2}{*}{NW-UCLA} \\ \cline{6-9}
                    &               &             &    &      & X-Sub(\%) & X-View(\%) & X-Sub(\%) & X-Set(\%) &      \\ \hline
MS-G3D \cite{liu2020disentangling}             & CVPR2020      & GCN         & 4S & 2.8 ($\downarrow$ 57.1\%) & 91.5      & 96.2      & 86.9      & 88.4      & -    \\
MST-GCN \cite{chen2021multi}            & AAAI2021      & GCN         & 4S & 12.0 ($\downarrow$ 90.0\%) & 91.5      & 96.6      & 87.5      & 88.8      & -    \\
CTR-GCN \cite{chen2021channel}             & ICCV2021      & Hybrid GCN  & 4S & 1.5 ($\downarrow$ 20.0\%) & 92.4      & 96.4      & 88.9      & 90.4      & 96.5 \\
EfficientGCN-B4 \cite{song2022constructing}    & TPAMI2022     & Hybrid GCN  & 4S & 2.0 ($\downarrow$ 40.0\%) & 91.7      & 95.7      & 88.3      & 89.1      & -    \\
InfoGCN \cite{chi2022infogcn}             & CVPR2022      & Hybrid GCN  & J  & 1.6 ($\downarrow$ 25.0\%) & 89.9      & 95.2      & 85.1      & 86.3      & -    \\
InfoGCN \cite{chi2022infogcn}            & CVPR2022      & Hybrid GCN  & 4S & 1.6 ($\downarrow$ 25.0\%) & 92.3      & 96.7      & 89.2      & 90.7      & 96.6 \\
FRHead \cite{zhou2023learning}             & CVPR2023      & Hybrid GCN  & J  & 2.0 ($\downarrow$ 40.0\%) & 90.3      & 95.3      & 85.5      & 87.3      & -    \\
FRHead \cite{zhou2023learning}             & CVPR2023      & Hybrid GCN  & 4S & 2.0 ($\downarrow$ 40.0\%) & 92.8      & 96.8      & 89.5      & 90.9      & 96.8 \\
HD-GCN \cite{Lee_2023_ICCV}             & ICCV2023      & Hybrid GCN  & J  & 1.7 ($\downarrow$ 29.4\%) & 93.0         & 97.1         & 89.8      & 91.2      & 96.9    \\
BlockGCN \cite{zhou2024blockgcn}              & CVPR2024      & Hybrid GCN  & J  & 1.3 ($\downarrow$ 7.7\%) & 90.9      & 95.4      & 86.9      & 88.2      & 95.5 \\
BlockGCN \cite{zhou2024blockgcn}            & CVPR2024      & Hybrid GCN  & 4S  & 1.3 ($\downarrow$ 7.7\%) & 93.1      & 97.0      & 90.3     & 91.5      & 96.9 \\ \hline
ST-TR \cite{plizzari2021skeleton}           & CVIU'21       & Transformer  & 4S  & 12.1 ($\downarrow$ 90.1\%)  & 90.3       & 96.3        & 85.1      & 87.1    & - \\
DSTA \cite{shi2020decoupled}                & ACCV'20      & Transformer   & 4S  & 4.1 ($\downarrow$ 70.7\%) & 91.5       & 96.4       & 86.6         & 89.0   & - \\
TranSkeleton \cite{liu2023transkeleton}         & TCSVT23       & Transformer & J  & 2.2 ($\downarrow$ 45.5\%) & 90.1      & 95.4      & 84.9      & 86.3      & -    \\
TranSkeleton \cite{liu2023transkeleton}        & TCSVT23       & Transformer & 4S & 2.2 ($\downarrow$ 45.5\%) & 92.8      & 97.0      & 89.4      & 90.5      & -    \\
Hyperformer \cite{zhou2022hypergraph}         & arXiv2022     & Transformer & 4S & 2.7 ($\downarrow$ 55.6\%) & 92.9      & 96.5      & 89.9      & 91.3      & 96.7 \\
SkeMixFormer \cite{xin2023skeleton}       & ACMM2023      & Transformer & J  & 2.1 ($\downarrow$ 42.9\%) & 90.7      & 95.9      & 87.1      & 88.9      & 96.8 \\
SkeMixFormer \cite{xin2023skeleton}        & ACMM2023      & Transformer & 4S & 2.1 ($\downarrow$ 42.9\%) & 92.8      & 96.9      & 90.0      & 91.2      & 97.1 \\\hline
FreqMixFormer \cite{wu2024frequency}       & ACMM2024      & Transformer & J  & 2.0 ($\downarrow$ 40.0\%)  & 91.5      & 96.0      & 87.9      & 89.1      & 96.8 \\
FreqMixFormer \cite{wu2024frequency}       & ACMM2024      & Transformer & 4S & 2.0 ($\downarrow$ 40.0\%)  & 93.4      & 97.3      & 90.2      & 91.5      & 97.4 \\ \hline
\rowcolor{lightgray}
\multicolumn{2}{c}{\textbf{FreqMixFormerV2}} & Transformer & J  & \textbf{\textcolor{red}{1.2}}  & 90.7      & 95.4         & 87.6          &  88.5         & 96.2     \\
\rowcolor{lightgray}
\multicolumn{2}{c}{\textbf{FreqMixFormerV2}} & Transformer & 4S & \textbf{\textcolor{red}{1.2} } & 92.9      & 96.9         & 90.0          &  91.1         & 97.0     \\ \hline
\end{tabular}
}
\label{tab: results}
\vspace{-5pt}
\end{table*}
\begin{figure}[ht]
\vspace{-5pt}
  \centering
  \includegraphics[width=1\linewidth]{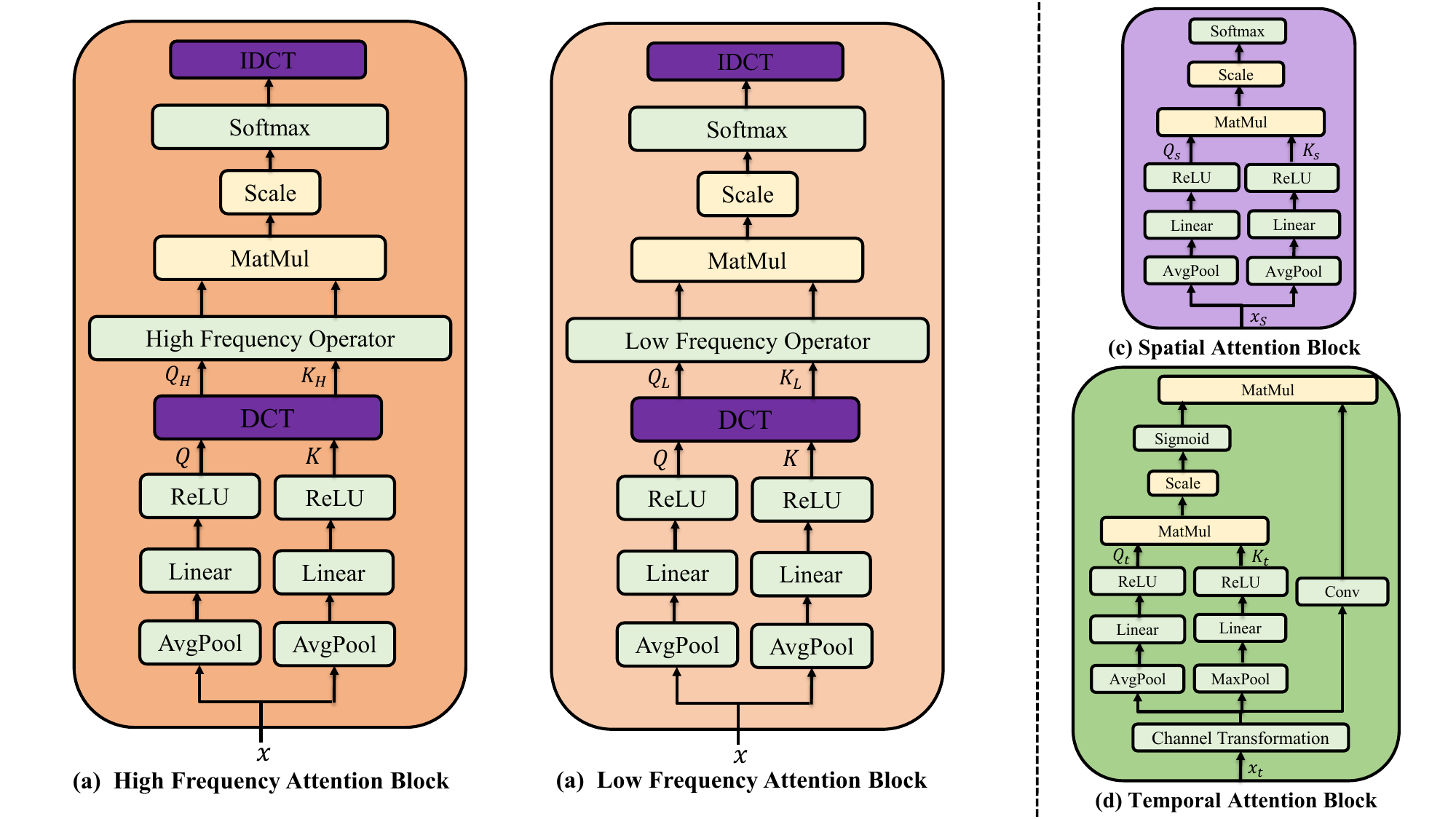}
  \vspace{-15pt}
  \caption{The mixed frequency blocks in FreqMixformerV2: (a) High-Frequency Attention Block (HFAB), (b) Low-Frequency Attention Block (LFAB). The common blocks applied in FreqMixFormer and FreqMixFormerV2 are (c) Spatial Attention Block and (d) Temporal Attention Block.}
  \label{fig:fig3}
  \vspace{-15pt}
\end{figure}
\subsubsection{Mixed Temporal Attention and Action Head}
Following \cite{wu2024frequency} and \cite{xin2023skeleton}, the spatial-frequency feature \(x_t\) is processed by a temporal attention block for temporal feature extraction:

 \vspace{-15pt}
\begin{align}
\label{eq:TEM}
\small 
       Q_t = \sigma({linear}({AvgPool}(CT(x_t)))), \\
       K_t = \sigma({linear}({MaxPool}(CT(x_t)))) 
\end{align} 
where channel transformation \( CT(\cdot) \) is a tricky method introduced in \cite{xin2023skeleton} for enhancing channel relationship learning for mix-attention mechanism. Then, the final output for the classification head is expressed as:

\vspace{-5mm}
\begin{align}
\label{eq:TemFinal}
\small 
     X_{out} = \left(\text{Sigmoid}\left(\text{Softmax}\left(\frac{Q_t K_t^\top}{\sqrt{d}}\right)\right)\right) V_t
\end{align}

Finally, an FC layer is adapted as the action head for classification. 

\section{EXPERIMENTS}
\subsection{Datasets and Implementation}


NTU RGB+D (NTU-60) \cite{shahroudy2016ntu}, NTU RGB+D 120 (NTU-120) \cite{liu2019ntu}, and Northwestern-UCLA (NW-UCLA) \cite{wang2014cross} are widely used action recognition datasets. NTU-60 and NTU-120 provide diverse skeletal action samples across various subjects and viewpoints, while NW-UCLA offers multi-view video clips for action recognition. The standard evaluation protocols \cite{wu2024frequency, xin2023skeleton, chi2022infogcn} are utilized for training and testing. The experiment setting is included in the \textcolor{magenta}{Appendix}.

\subsection{ Comparison with the State-of-the-Art}
Across large-size datasets, FreqMixFormerV2 consistently demonstrates superior efficiency while maintaining competitive accuracy. On the NTU-60 dataset, our model achieves 90.7\% accuracy on the X-Sub setting and 92.9\% on the X-View setting, all with a model size of only 1.2M parameters. This represents a 40.0 \% reduction in parameter count compared to the original FreqMixFormer, yet FreqMixFormerV2 continues to deliver state-of-the-art performance. The efficiency of FreqMixFormerV2 is further highlighted on the NTU-120 dataset, where it attains an accuracy of 87.6\% (X-Sub) and 90.0\% (X-Set). These results are achieved with the same reduced model size, substantially improving computational efficiency over many existing models without compromising performance. FreqMixFormerV2 also achieves an impressive accuracy of 96.2\% on the small-size NW-UCLA dataset, further underscoring its efficiency. Notably, more experiments on confusing action are listed in the \textcolor{magenta}{Appendix}.
\begin{table}[]
\renewcommand\arraystretch{1.2}
\centering
\caption{Search for the Best Partition $N$ of High and Low-Frequency Coefficients and Best High-Frequency Operator $h$ and Low-Frequency Operator $\ell$.}
\setlength\tabcolsep{6.0pt} 
\label{tab:combined}
\begin{tabular}{ccc|llc}
\hline
\multicolumn{3}{c|}{\textbf{Search for best $N$ }} & \multicolumn{3}{c}{\textbf{Search for best $h$ and $\ell$}} \\ 
\hline
$N$ & NTU-60 X-Sub (\%) &  & $\ell$ & $h$ & NTU-60 X-Sub (\%) \\ \hline
1 & 90.2 &  & 0.1 & 1.1 & 90.0 \\
3 & 90.3 &  & 0.2 & 1.2 & \textbf{90.7} \\
5 & 90.4 &  & 0.3 & 1.3 & 90.1 \\
7 & 90.3 &  & 0.4 & 1.4 & 90.2 \\
9 & 90.4 &  & 0.5 & 1.5 & 90.4 \\
11 & 90.5 &  & 0.6 & 1.6 & 90.0 \\
13 & \textbf{90.7} &  & 0.7 & 1.7 & 90.0 \\
15 & 90.6 &  & 0.8 & 1.8 & 90.0 \\
17 & 90.2 &  & 0.9 & 1.9 & 90.1 \\ \hline
\end{tabular}
\label{tab:ablation12}
\end{table}
\vspace{-10pt}
\subsection{Ablation Study}
\subsubsection{Search for the Best Partition $N$ of High and Low Frequency Coefficients}

In Table \ref{tab:ablation12}, we evaluate the effect of different portions $N$, where \(N \in (1, 25)\) because there are 25 joints in each subject in the NTU-60 dataset. As we discussed in Section \ref{sec:high-low}, the first $N$ DCT coefficients represent low-frequency components and the remaining coefficients represent high-frequency components. As shown in Table. \ref{tab:ablation12}, partition $N$ = 13 achieves the best accuracy of 90.7\% on NTU-60 X-Sub. The experiments highlight that the optimal choice of $N$ ensures a balanced representation of frequency components, leading to improved recognition accuracy (0.5\%).
\subsubsection{Search for the Best High-Frequency Operator $h$ and Low-Frequency Operator $\ell$}
In Table \ref{tab:ablation12}, we also explore the effect of different combinations of the high-frequency operator \(h\) and low-frequency operator \(\ell\) on the NTU-60 X-Sub dataset. The parameter \(\ell\) controls the adjustment of low-frequency coefficients, while \(h\) adjusts the high-frequency coefficients. The results show that the combination of \(\ell = 0.2\) and \(h = 1.2\) achieves the highest accuracy of 90.7\%. This indicates that fine-tuning the frequency coefficients using the optimal \(h\) and \(\ell\) values is crucial for enhancing the model's ability to capture the most relevant features, leading to improved recognition accuracy.



\section{Conclusion}
This paper presents a Lightweight Frequency-aware Mixed Transformer (FreqMixFormerV2) for skeleton action recognition. By redesigning the attention blocks and introducing a new high-low-frequency operator, we have significantly reduced the model's parameter count and computational intensity while retaining its ability. Extensive experiments show that our proposed method successfully achieves a superior balance between efficiency and accuracy.

{\small
\bibliographystyle{ieee}
\bibliography{egbib}
}

\clearpage
\appendix
\textbf{Experiment Settings:}
We preprocess the skeleton data following the standard method from \cite{chen2021channel},\cite{chi2022infogcn},\cite{xin2023skeleton},\cite{wu2024frequency}. The proposed approach is implemented using PyTorch \cite{paszke2019pytorch} and trained on two NVIDIA RTX A6000 GPUs. The model is trained for 100 epochs with a batch size of 128 across all datasets, with a warm-up phase during the first 5 epochs. The weight decay is set to 0.0005. The initial learning rate is 0.1 for the NTU-60 and NTU-120 datasets, with reductions by a factor of 0.1 at the 35th, 55th, and 75th epochs. For the Northwestern-UCLA dataset, the learning rate is initialized to 0.2, with a 0.1 reduction at the 50th epoch. We also implement a multi-stream ensemble method \cite{chi2022infogcn}, \cite{xin2023skeleton} for 4-stream fusion. 

\textbf{Discrete Cosine Transform (DCT) for Joint Encoding:}
We follow the same method as FreqMixFormer\cite{wu2024frequency} to introduce DCT for joint sequence encoding: 
Let \( x \in \mathbb{R}^{J \times C \times F} \) denote the input joint sequence, where the trajectory of the \( j \)-th joint across \( F \) frames is represented as \( X_j = (x_{j,0}, x_{j,1}, \dots, x_{j,F}) \). While traditional transformer-based skeleton action recognition methods use \( X_j \) as input for spatial domain correlation learning, we propose leveraging a frequency representation via the Discrete Cosine Transform (DCT). Our approach retains all DCT coefficients, enhancing the high-frequency parts and reducing the low-frequency parts using high-frequency operator $h$ and low-frequency operator $\ell$.

We apply DCT to each joint trajectory \( X_j \), with the \( i \)-th DCT coefficient calculated as:
\begin{align}
\label{eq:dct}
\small
C_{j,i} = \sqrt{\frac{2}{F}} \sum_{f=1}^{F} x_{j,f} \frac{1}{\sqrt{1 + \delta_{i1}}} \cos\left[\frac{\pi(2f - 1)(i - 1)}{2F}\right],
\end{align}
where \( \delta_{ij} \) is the Kronecker delta. Here, larger \( i \) values correspond to higher frequency components. The original time-domain sequence can be restored using the Inverse Discrete Cosine Transform (IDCT):
\begin{align}
\label{eq:idct}
\small
x_{j,f} = \sqrt{\frac{2}{F}} \sum_{i=1}^{F} C_{j,i} \frac{1}{\sqrt{1 + \delta_{i1}}} \cos\left[\frac{\pi(2f - 1)(i - 1)}{2F}\right]
\end{align}

\textbf{The Design of Lightweight Frequency-aware Mixed Transformer:}
As shown in Table \ref{tab:ablation3}, the baseline model only includes the fundamental FreqMixFormer\cite{wu2024frequency}, achieving an accuracy of 88.5\% on the NTU-60 X-Sub benchmark. To improve upon this, we analyze three key modules: 1) High-Frequency Attention Block (HFAB): This module enhances the model's ability to capture subtle, high-frequency movements by focusing on the high-frequency components in the skeletal data. Incorporating HFAB into the baseline improves the accuracy by 0.7\%, reaching 89.2\%. 2) Low-Frequency Attention Block (LFAB): This module targets steady or static low-frequency movements. Adding LFAB results in a 1.6\% improvement, bringing the accuracy to 90.1\%. 3) Temporal Attention Block (TAB): This module is designed to learn correlations across frames, improving the temporal modeling of joint movements. Integrating TAB with the baseline leads to a 1.7\% increase in accuracy, achieving 90.2\%. When these modules are combined, the model achieves an accuracy of 90.7\%, a substantial 2.2\% improvement over the baseline. The significant enhancement comes from the comprehensive integration of HFAB, LFAB, and TAB, demonstrating that the lightweight design of the Frequency-aware Mixed Transformer is highly effective in improving action recognition performance.

\begin{table}[]
\renewcommand\arraystretch{1.0} 
\centering
\caption{The design of Lightweight Frequency-aware Mixed Transformer.}
\setlength\tabcolsep{6.0pt} 
\label{tab: design}
\begin{tabular}{ccccc}
\hline
\textbf{Baseline} & \textbf{HFAB} & \textbf{LFAB} & \textbf{TAB} & \textbf{NTU-60 X-Sub (\%)} \\ \hline
\Checkmark & \XSolidBrush & \XSolidBrush & \XSolidBrush & 88.5 \\
\Checkmark & \XSolidBrush & \XSolidBrush & \Checkmark & 89.2 ($\uparrow$ 0.7)\\
\Checkmark & \Checkmark & \XSolidBrush & \XSolidBrush & 90.1 ($\uparrow$ 1.6)\\
\Checkmark & \XSolidBrush & \Checkmark & \XSolidBrush & 90.2 ($\uparrow$ 1.7)\\
\Checkmark & \Checkmark & \Checkmark & \XSolidBrush & 89.4 ($\uparrow$ 0.9)\\
\Checkmark & \Checkmark & \Checkmark & \Checkmark & \textbf{90.7} ($\uparrow$ \textbf{2.2})\\ \hline
\end{tabular}
\label{tab:ablation3}
\end{table}

\begin{table}[]
\renewcommand{\arraystretch}{1.0} 
\centering
\setlength{\tabcolsep}{10pt} 
\caption{The results on different difficult-level actions for NTU RGB+D dataset.}
\vspace{-10pt}
\begin{tabular}{lcc} 
\hline
\multirow{2}{*}{Method} & \multicolumn{2}{c}{NTU60 X-Sub (\%)} \\ \cline{2-3} 
                        & Hard             & Medium           \\ \hline
Hyperformer            & 71.4             & 83.6             \\
SkeMixFormer           & 71.9             & 84.6             \\
FreqMixFormerV2        & 72.5             & 84.7             \\ \hline
\end{tabular}
\label{tab:diff}
\vspace{-10pt}
\end{table}

\begin{figure*}[]
\vspace{-10pt}
  \centering
  \includegraphics[width=0.98\linewidth]{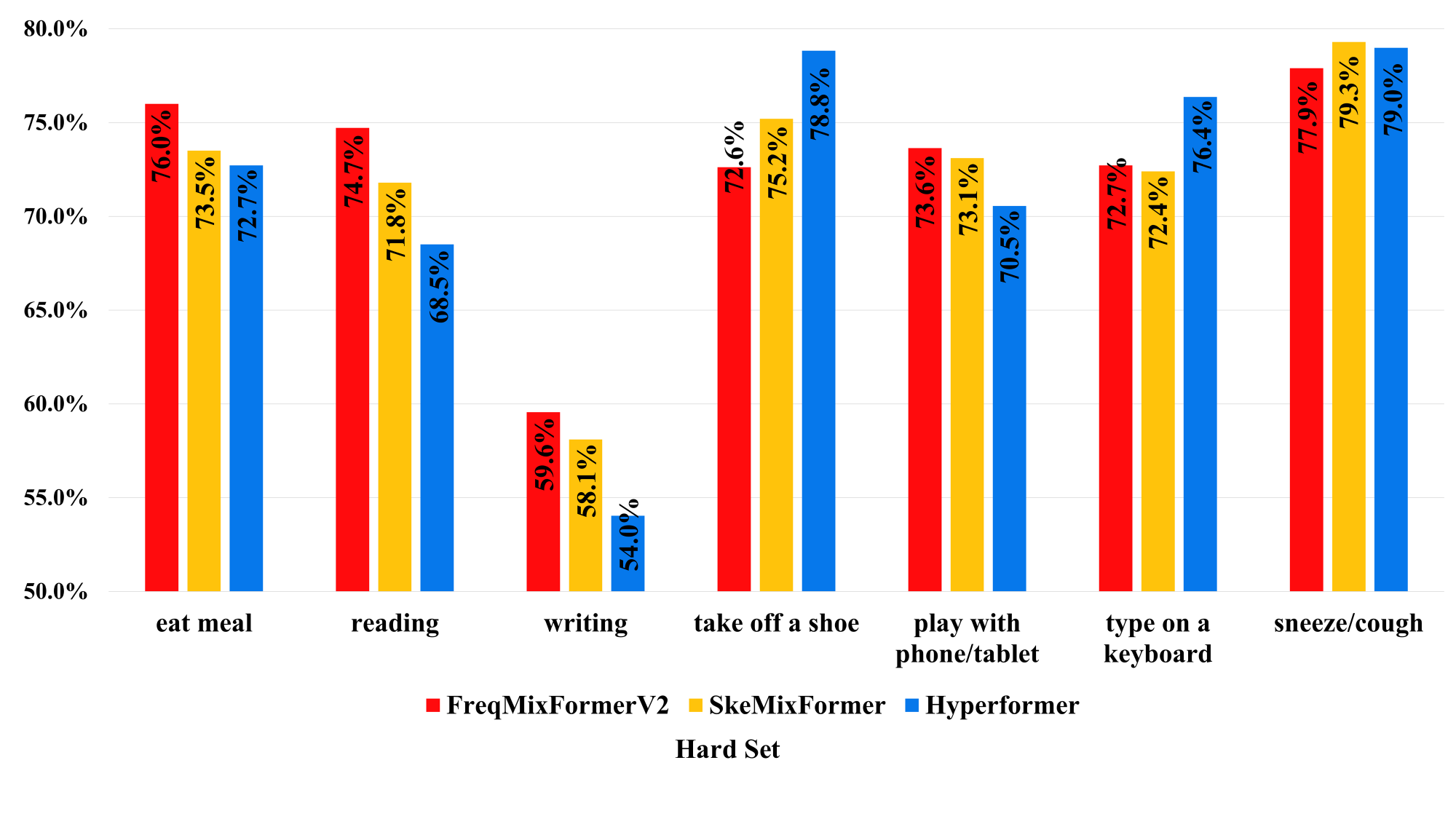}
  \vspace{-15pt}
  \caption{Accuracy comparison results on confusing actions in the hard set.}
  \label{fig:fig4}
  \vspace{-5pt}
\end{figure*}

\begin{figure*}[]
\vspace{-10pt}
  \centering
  \includegraphics[width=0.98\linewidth]{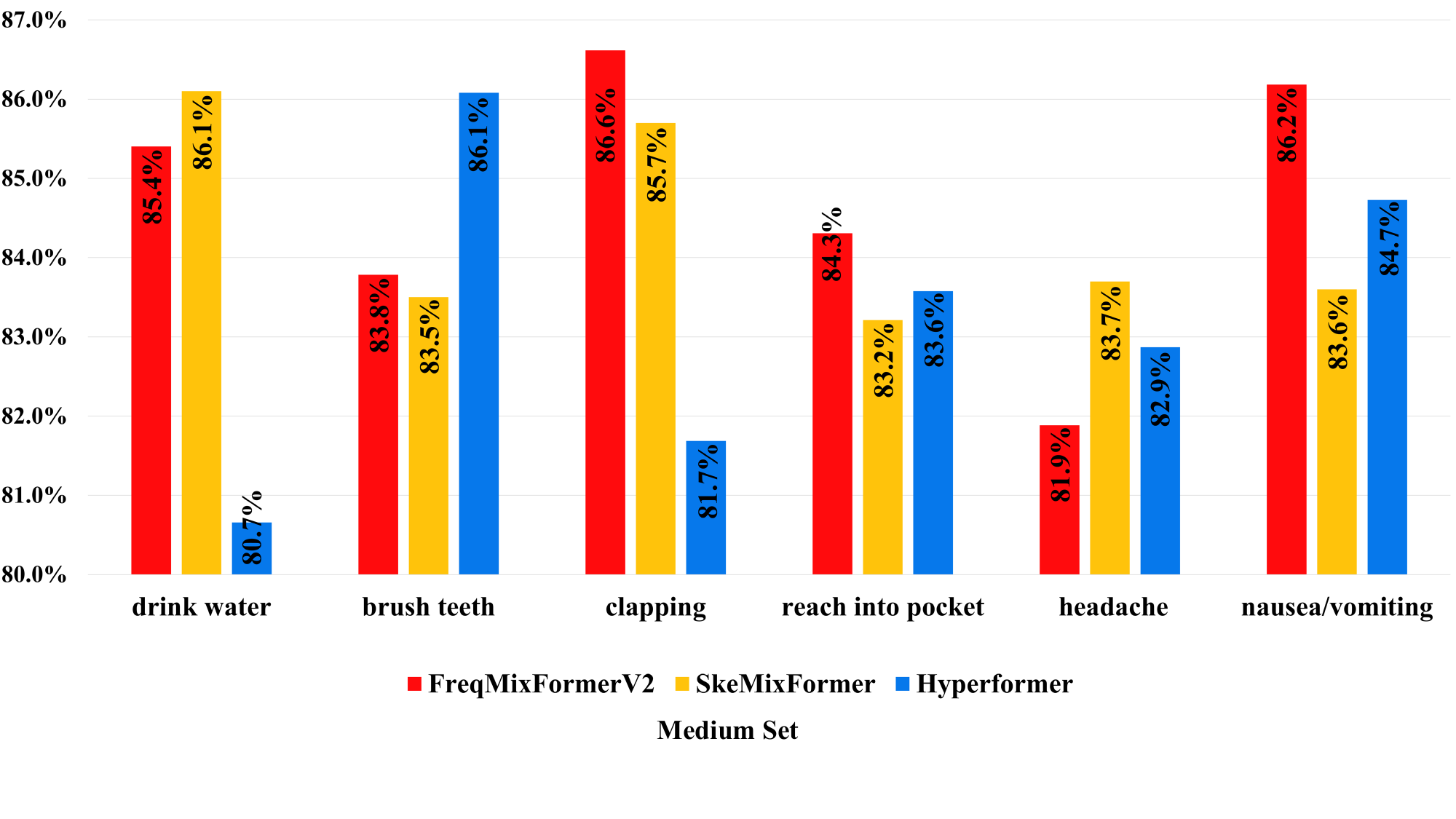}
  \vspace{-15pt}
  \caption{Accuracy comparison results on confusing actions in the medium set.}
  \label{fig:fig5}
  \vspace{-15pt}
\end{figure*}
\textbf{More Results on Confusing Action:}
We adopt the comparison method used in FreqMixFormer \cite{wu2024frequency} to evaluate our model's performance in recognizing subtle, confusing actions under the NTU-60 X-Sub setting (with joint modality only). For validation, we categorize the actions into two sets: the Hard set (actions with accuracy below 80\%) and the Medium set (actions with accuracy between 80\% and 90\%). We compare our results with transformer-based models SkeMixFormer \cite{xin2023skeleton} and HyperFormer \cite{zhou2022hypergraph}. The results are shown in Fig. \ref{fig:fig4} and Fig. \ref{fig:fig5}. Despite utilizing significantly fewer parameters than previous methods, our model surpasses the others in recognizing the most challenging actions with subtle movements.

\end{document}